\documentclass[letterpaper]{article} 
\usepackage[preprint]{aaai2027}

\usepackage[hyphens]{url}  
\usepackage{graphicx} 
\usepackage{natbib}   
\usepackage{caption}  
\usepackage{algorithm}
\usepackage{algorithmic}
\usepackage{amsmath}
\usepackage{amssymb}

\usepackage{newfloat}
\usepackage{listings}
\DeclareCaptionStyle{ruled}{labelfont=normalfont,labelsep=colon,strut=off} 
\floatstyle{ruled}
\newfloat{listing}{tb}{lst}{}
\floatname{listing}{Listing}

\usepackage{booktabs}

\title{Connectome-to-Function: Conditional Generative Latent Representations for Reservoir Computing}
\author{
Zhuolin Yu\textsuperscript{\rm 1, 2}\equalcontrib,
Xingyu Liu\textsuperscript{\rm 1}\equalcontrib,
Yuanhao Jia\textsuperscript{\rm 1,\rm 3},
Yunhang Xiao\textsuperscript{\rm 4},
Hairuo Xue\textsuperscript{\rm 5},
Feihan Sun\textsuperscript{\rm 1, 2},
Guozhang Chen\textsuperscript{\rm 1}\corresponding
}

\affiliations{
\textsuperscript{\rm 1}
National Key Laboratory for Multimedia Information Processing, School of Computer Science, Peking University, Beijing, China\\
\textsuperscript{\rm 2}
School of Electronics Engineering and Computer Science, Peking University, Beijing, China\\
\textsuperscript{\rm 3}
School of Computer Science, Beijing University of Posts and Telecommunications, Beijing, China\\
\textsuperscript{\rm 4}
College of Engineering, Peking University, Beijing, China\\
\textsuperscript{\rm 5}
Yuanpei College, Peking University, Beijing, China\\
guozhang.chen@pku.edu.cn
}

\begin{document}

\maketitle

\begin{abstract}
Connectomes, graph-level maps of neurons and their synaptic connections, provide a structural basis for understanding how brain circuits support function and computation. However, mapping connectome structure to computation remains difficult because these graphs are high-dimensional, sparse, and sensitive to local structural variation. Existing approaches often depend on hand-crafted structural descriptors or task-specific predictors, which limits their ability to represent connectomes in a form that is both generative and functionally meaningful. We propose a conditional generative latent framework that encodes connectome graphs into a compact structural space while using available node-level conditions to guide reconstruction and generation. From this space, the model can reconstruct observed connectivity with a mean edge-reconstruction AUC up to 0.910 and generate new candidate connectomes, enabling a unified analysis of graph structure and computational behavior. Using connectome-derived graphs as recurrent computational substrates, we found that the learned latent space captures functional variation across reservoir-computing experiments, with cross-validated $R^2$ values up to approximately 0.87. Interpretability analysis further revealed task-specific structural mechanisms: in our examples, memory performance is associated with reciprocal recurrent connectivity, whereas prediction and classification are more strongly associated with spectral properties of the recurrent network. These findings suggest an AI-for-science approach to linking neural connectivity to computation and provide a generative and interpretable basis for studying how distinct structural mechanisms shape computational capacity.
\end{abstract}

\section{Introduction}

A central scientific question in neuroscience is how the structure of a neural circuit gives rise to its computational function. Connectomes provide increasingly detailed descriptions of neural wiring across scales \cite{sporns2005human}, and network neuroscience has related topology to integration, segregation, and wiring cost \cite{bullmore2009complex,rubinov2010complex}. Recent single-cell connectomes further resolve synaptic connectivity together with neuronal 3D position and cell identity \cite{dorkenwald2024neuronal,microns2025functional}, making this structure--function question accessible at the level of local circuits. Reservoir computing provides a controlled way to probe such relationships because recurrent connectivity is kept fixed while only a readout is trained \cite{jaeger2001echo,maass2002real,suarez2024connectome}. This leads to our central question: \emph{can measured connectome structure be organized into a continuous representation that helps explain variation in circuit computation?}

\begin{figure}[t]
\centering
\includegraphics[width=\columnwidth]{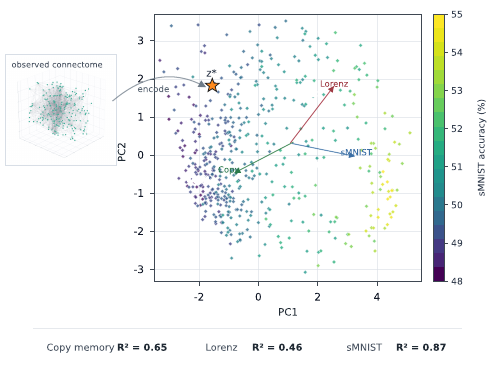}
\caption{Connectome-to-function latent space. Observed connectomes are encoded into a generative latent space where coordinates predict reservoir performance across copy memory, Lorenz system prediction, and sequential MNIST. Arrows indicate task-specific functional directions, and bottom values report cross-validated $R^2$.}
\label{fig:teaser}
\end{figure}

Existing approaches capture only parts of this problem. Graph descriptors characterize empirical connectomes through predefined statistics \cite{sporns2005human,bullmore2009complex,rubinov2010complex}, while generative network can synthesize graph structure \cite{vertes2012simple,betzel2016generative,betzel2017generative,akarca2021generative,barabasi2020genetic,kipf2016variational,simonovsky2018graphvae}; however, these approaches are primarily designed to reproduce structural properties rather than to reveal how structural variation maps to computation. Shuvaev et al. showed that a compact generative encoding can preserve task-relevant network structure, evaluating compression by the performance of the generated network rather than weight reconstruction \cite{shuvaev2024encoding,gaier2019weight}. However, their representation is obtained for task-trained artificial networks with a specific task performance entering the optimization criterion. Conversely, connectome-based reservoir studies directly evaluate biological wiring as a computational substrate \cite{damicelli2022brain,morra2023using,costi2025drosophila,sumi2023biological}, but largely study fixed connectomes or rewired variants. Thus, a key gap remains: a \emph{generative representation learned solely from measured biological connectivity} in which computational organization can be discovered without using function to train the representation.

To address this gap, we propose a conditional generative latent framework for sparse connectome graphs \cite{sohn2015learning}. The model encodes each local circuit into a compact structural coordinate while using neuronal location and cell type as generation conditions, enabling both reconstruction and controlled sampling of candidate connectomes. Experiments show that this latent space is structurally faithful and functionally informative. The model reaches an edge-reconstruction AUC of $0.910$ and better preserves higher-order topology than a naive VAE baseline \cite{liu2026decodingcorticalmicrocircuitsgenerative}. When generated graphs are evaluated as reservoirs on memory, forecasting, and classification tasks \cite{suarez2024connectome}, their latent coordinates predict performance with cross-validated $R^2$ values of approximately $0.46$--$0.87$. Interpretability analysis further reveals task-specific structural mechanisms: memory is associated with reciprocal recurrent connectivity, whereas prediction and classification are more strongly associated with spectral properties of the recurrent network. Thus, the learned coordinates provide a generative graph representation aligned with measurable computational variation.

\begin{figure*}[t]
\centering
\includegraphics[width=\textwidth]{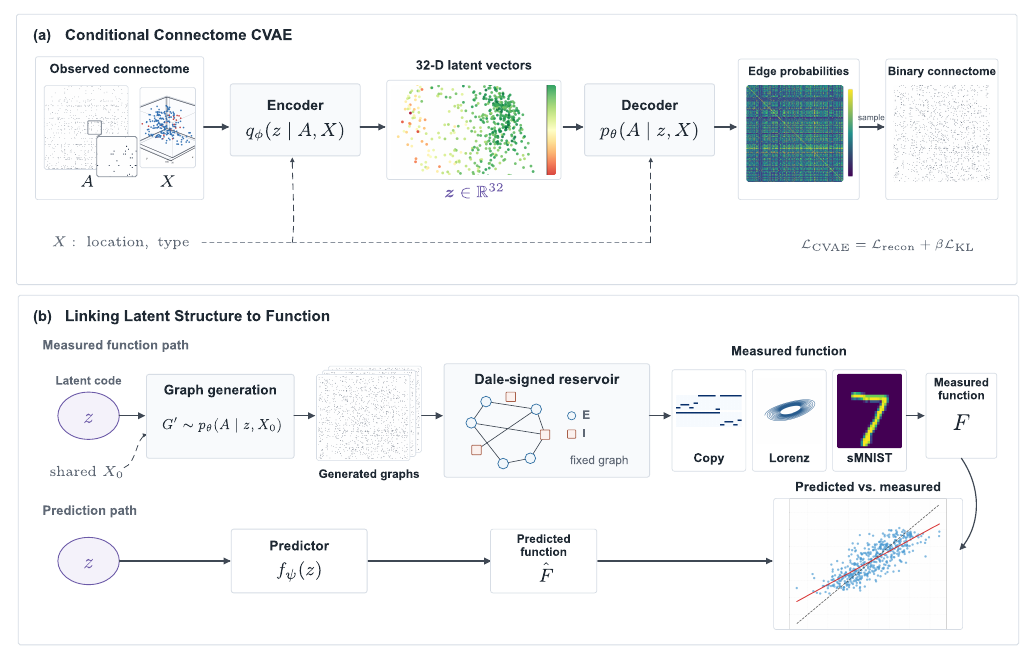}
\caption{Framework overview. (a) The conditional VAE encodes directed local circuits sampled from connectomes with node coordinates and cell type into a 32-dimensional structural latent coordinate, then decodes edge probabilities under node-level conditions. (b) Decoded graphs define fixed Dale-signed reservoirs evaluated on copy memory, Lorenz system prediction, and sequential MNIST; a predictor maps latent coordinates $z_i$ to functional scores $F_i$.}
\label{fig:overall-framework}
\end{figure*}

In summary, our main contributions are as follows: 
\begin{itemize}
    \item We propose a conditional generative framework for connectome graphs that learns latent representations by conditioning on neuronal cell type and spatial organization.
    \item We identify a structure-function space predictive of reservoir-computing performance in the learned latents, across various tasks including memory, dynamic prediction, and classification.
    \item We conduct an interpretability analysis of the learned connectome space, linking task-relevant functional directions to distinct structural mechanisms. 
\end{itemize}

\section{Related Work}

\paragraph{Graph Descriptor and Generative Representation Learning}
Connectomes have traditionally been characterized using graph-theoretic descriptors that summarize properties such as small-worldness, modularity, efficiency, clustering, and wiring cost \cite{sporns2005human,bullmore2009complex,rubinov2010complex}. While these measures provide interpretable summaries of network organization, they are predefined, post-hoc descriptive statistics and therefore offer limited access to the high-dimensional space of possible connectome structures. Mechanistic generative network models address this limitation by synthesizing graphs under prescribed wiring rules and structural constraints \cite{vertes2012simple,betzel2016generative,betzel2017generative,akarca2021generative,barabasi2020genetic}, yet they lack continuous, data-driven latent spaces. In parallel, graph representation learning provides continuous representations of discrete graph structures: variational graph autoencoders and related models encode graphs or their nodes into latent variables to reconstruct adjacency patterns \cite{kipf2016variational,simonovsky2018graphvae}, while conditional variational frameworks enable generation under external attributes \cite{sohn2015learning}. Although models such as GraphVAE can generate small graphs under graph-size constraints \cite{simonovsky2018graphvae}, general-purpose graph generators remain difficult to apply directly to neural microcircuits. Common decoding or graph-matching schemes are not well suited to the extreme sparsity of cortical connectivity, and existing models rarely integrate 3D spatial coordinates, discrete cell identities, and permutation-invariant graph-level representation within a single conditional generative process.

\paragraph{Connectome-Based Computation and Reservoir Models}

Recent network-neuroscience studies have leveraged empirical connectomes as recurrent architectures, often using reservoir-computing models to relate static topology to computation \cite{suarez2024connectome}. Prior work has shown that connectome-derived reservoirs can support memory tasks \cite{damicelli2022brain}, chaotic time-series prediction under fruit-fly-connectome constraints \cite{morra2023using}, and Drosophila-based forecasting analyses involving empirical topology and weight distributions \cite{costi2025drosophila}. Related biological reservoir studies also suggest that modular neuronal organization can improve generalization-filter behavior \cite{sumi2023biological}. Although these studies show that empirical connectomes can support reservoir-computing tasks, they mainly evaluate fixed empirical networks or simple rewired variants. They do not learn a continuous generative latent space that can both produce new connectome structures and predict functional variation across tasks.

\paragraph{Structure-to-Function Mapping and Performance Prediction}

Investigating how network topology dictates computational capacity forms a shared frontier across machine learning and systems neuroscience. Neural architecture search (NAS) studies typically analyze artificial architectures and use architecture--performance pairs or initialization-based proxies to estimate task performance \cite{you2020graph,wen2020neural,abdelfattah2021zero}. In systems neuroscience, connectome-constrained models show that measured biological wiring can support mechanistic predictions for a specified computation \cite{lappalainen2024connectome}, while spatially embedded recurrent networks demonstrate how structural and functional organization can emerge under biological constraints in artificial systems \cite{achterberg2023spatially}. Across these lines of work, structure--function relationships are typically studied either by evaluating task-trained artificial architectures, by using task performance to supervise architecture prediction, or by optimizing models constrained by a particular biological circuit. Here, we take a complementary approach: we learn a continuous generative representation from measured biological connectivity without task-performance supervision, and subsequently test whether the learned space predicts variation in reservoir performance across tasks.

\section{Method}

\subsection{Conditional VAE for Connectome}

\paragraph{Problem Setup and Notation}
Single-cell-level structure is high dimensional.
To better address the structure-to-function question, we first learn a low-dimensional latent space for circuit structure. Specifically, a local circuit structure is considered as an attributed directed graph $ G = (V, E, X)$, where $V$ and $E$ are the sets of vertices and edges and $X$ denotes the node-annotation matrix with $N = |V|$. Further, we used adjacency matrix $A \in \{0,1\}^{N \times N}$, where $A_{ij}=1$ indicates a connection from neuron $j$ to neuron $i$. We decompose the node annotations as 
\[
X = \operatorname{Concat}\left(X_{\mathrm{loc}}, X_{\mathrm{type}}\right)
\]
where $X_{\mathrm{loc}} \in \mathbb{R}^{N \times 3}$ contains the spatial coordinates of the neuronal somas and $X_{\mathrm{type}} \in \{0, 1 \}^{N \times N_{ \mathrm{type}}}$ encodes neuron types (excitation or inhibition) as one-hot vectors. 
We parameterize this latent space by assigning each graph a $D$-dimensional latent variable $z \in \mathbb{R}^{D}$.
To this end, the encoder, parameterized by $\phi$, defines the approximate posterior distribution $q_{\phi}(z \mid A, X)$, while the decoder, parameterized by $\theta$, defines the conditional likelihood $p_{\theta}(A \mid z, X)$.
Thus, the latent variable $z$ captures the structural information of the connectome graph conditioned on the observed node features. 

\paragraph{Conditional Variational Objective}
Conditioning on the node features $X$, we formulate connectome structure modeling under a conditional variational framework \cite{kingma2014autoencoding,rezende2014stochastic}. Specifically, we assume a standard Gaussian prior over $z$ that is independent of the node conditions $X$, i.e., $p(z \mid X)=p(z)=\mathcal{N}(0,I)$. Under this generative assumption, the conditional log-likelihood $\log p_{\theta}(A \mid X)$ admits the following evidence lower bound (ELBO):

$$
\begin{aligned}
\log p_{\theta}(A \mid X)
&\geq \mathcal{L}_{\mathrm{ELBO}}(\theta,\phi;A,X) \\
&= \mathbb{E}_{q_{\phi}(z \mid A,X)}
\left[\log p_{\theta}(A \mid z,X)\right] \\
&\quad - D_{\mathrm{KL}}
\left(q_{\phi}(z \mid A,X)\,\|\,p(z)\right).
\end{aligned}
$$

For training, we introduce a coefficient $\beta$ to control the strength of the latent regularization, following the beta-VAE formulation \cite{higgins2017beta}. The resulting objective is

$$
\mathcal{L}_{\mathrm{CVAE}}
=
\mathcal{L}_{\mathrm{recon}}
+
\beta \mathcal{L}_{\mathrm{KL}},
$$

where

$$
\mathcal{L}_{\mathrm{recon}}
=
-\mathbb{E}_{q_{\phi}(z \mid A,X)}
\left[\log p_{\theta}(A \mid z,X)\right],
$$
$$
\mathcal{L}_{\mathrm{KL}}
=
D_{\mathrm{KL}}
\left(q_{\phi}(z \mid A,X)\,\|\,p(z)\right).
$$

\paragraph{Architecture}
We formulate the problem as conditional connectome graph generation, where the graph is generated conditioned on node-level information. The connectome data exhibit two salient properties: first, the 3D neuron coordinates provide explicit spatial structure; second, the graph representation and the learned latent embedding should be invariant to node permutations. To account for both properties, we introduce PointNet++ as an essential encoding layer on the encoder side, where it hierarchically fuses node attributes with spatial coordinates into local geometric embeddings \cite{qi2017pointnetplusplus}. These embeddings are then processed by a GAT to integrate adjacency information \cite{velickovic2018graph}, and by a Transformer-based graph global encoder with a learned CLS token to obtain a permutation-invariant graph-level representation \cite{vaswani2017attention}. Two projection heads finally parameterize $\mu$ and $\log \sigma^{2}$, from which the latent variable is sampled via the standard reparameterization trick \cite{kingma2014autoencoding,rezende2014stochastic}. 

On the decoder side, we follow the same design principle. The conditional template is first encoded by PointNet++ to produce node-wise condition embeddings, and a Transformer decoder combines these embeddings with the latent code to reconstruct node representations. An edge predictor then maps the decoded node representations to a directed edge-probability matrix, on which Bernoulli sampling is performed to obtain binary adjacency matrices for downstream tasks. In this way, our decoder preserves the conditional generation setting of the reference model, while making the spatial structure of the input explicit throughout decoding. Please see Supplementary Materials for details.

\subsection{Linking Latent Structure to Function}
\paragraph{Connectome-based Reservoir Computing}
To evaluate the functional properties of each connectome graph, we instantiate it as a recurrent reservoir and measure its computational performance across tasks. To follow Dale's rule \cite{dale1935pharmacology}, we assign a sign $d_{j} \in \{-1, 1\}$ to presynaptic neuron $j$ according to its inhibitory or excitatory type, respectively, and construct the recurrent matrix as $W_{ij} = A_{ij} d_{j}$. To make reservoir dynamics comparable across graphs, we normalize the recurrent matrix to a fixed spectral radius before task evaluation.

We use the resulting matrix $W$ as the recurrent connectivity in the standard reservoir-computing model \cite{jaeger2001echo,maass2002real,lukosevicius2009reservoir}. Given a sequence input $\{x_{t}\}_{t=1}^T$, the computation consists of input injection, recurrent state updates, and linear readout:
\[
\begin{aligned}
a_t &= W_{\mathrm{in}}x_t + Wh_{t-1},\\
h_t &= (1-\eta)h_{t-1} + \eta\tanh(a_t),\\
\hat{y}_t &= W_{\mathrm{out}}h_t.
\end{aligned}
\]
Here, $h_{t}$ denotes the reservoir state, $\eta$ is the leaking rate, $W_{\mathrm{in}}$ maps the input to the reservoir state space, and $W_{\mathrm{out}}$ maps the state to the task output.

During training, $W_{\mathrm{in}}$ is randomly initialized and then held fixed, as is $W$. Only $W_{\mathrm{out}}$ is fitted to the task targets, yielding $W_{\mathrm{out}}^{\star}$. Nothing changes in testing phase. Thus, performance between local circuits measures how recurrent topology affects reservoir dynamics and the information available to a linear output.

\paragraph{Latent-to-Function Regression}
Given the connectome graphs $\{G_i\}_{i=1}^{M}$, we first encode each graph with the CVAE encoder to obtain latent vectors $\{z_i\}_{i=1}^{M}$, where each $z_i$ summarizes the structural pattern of the corresponding local circuit. We then examine whether differences across these structural latent representations predict variation in computational function by decoding each latent code into a connectome and evaluating the resulting graph as a reservoir.

Because both the latent code $z_i$ and the node-level conditions $X_i$ can influence the decoded connectivity, we fix the node conditions across latent codes to isolate structural variation attributable to $z$. Specifically, we randomly select a local circuit as a shared condition template, denoted by $X_0$. We repeat the analysis using 10 different templates and observe no qualitative change in the results. Each latent code $z_i$ is then decoded under the same condition $X_0$:

$$
G_i' \sim p_{\theta}(A \mid z_i, X_0).
$$

The function of each generated graph is evaluated using the reservoir-computing pipeline. Because both graph generation and reservoir evaluation involve stochasticity, we estimate a robust functional score for each latent code. Specifically, for each $z_i$, we generate 10 binary adjacency matrices under the shared condition $X_0$, evaluate each graph across 3 random seeds, and average the resulting task scores. This yields one aggregated functional score $F_i$ for each latent code $z_i$.

We then fit a regression model from the latent representation to the predicted functional score $\hat{F}_i$,
$$
f_{\psi}: z_i \rightarrow F_i,
\qquad
\hat{F}_i = f_{\psi}(z_i).
$$
Reliable cross-validated prediction indicates that the learned latent representation contains structural variation that is informative of computational function.

\begin{figure*}[t]
    \centering
    \includegraphics[width=\textwidth]{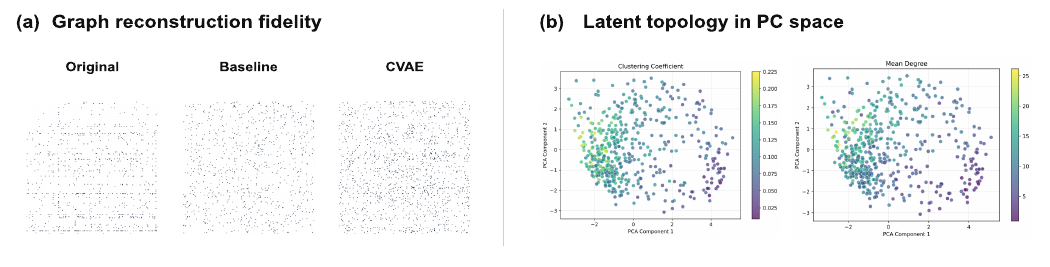}
    \caption{Connectome reconstruction fidelity and latent topological organization. (a) Representative binary adjacency matrices for the original connectome, the naive VAE baseline, and the full CVAE reconstruction. (b) PCA projection of the learned 32-dimensional latent vectors from 498 connectome samples, colored by clustering coefficient and mean degree. The smooth variation of these graph-level descriptors over the PC1--PC2 plane indicates that the latent space preserves continuous macroscopic topological variation.}
    \label{fig:reconstruction_latent_topology}
\end{figure*}

\begin{table*}[t]
\centering
\begin{tabular}{@{}lccccccccc@{}}
  \toprule
  Model & AUC $\uparrow$ & Deg. $\downarrow$ & Eff. $\downarrow$ & Clust. $\downarrow$ & Assort. $\downarrow$ & Mod. $\downarrow$ & Trans. $\downarrow$ & Louv. $\downarrow$ \\
  \midrule
  Naive VAE baseline & 0.722  & \textbf{0.084} & \textbf{0.288} & 0.820 & 1.681 & 0.326 & 0.733 & 0.392 \\
  Full CVAE & \textbf{0.910}  & 0.198 & 0.314 & \textbf{0.380} & \textbf{1.210} & \textbf{0.090} & \textbf{0.163} & \textbf{0.119} \\
  \bottomrule
\end{tabular}
\caption{Reconstruction comparison between the full CVAE and naive VAE baseline. AUC measures edge-level reconstruction. Deg., Eff., Clust., Assort., Mod., Trans., and Louv. denote mean degree, global efficiency, clustering coefficient, assortativity, modularity, transitivity, and directed Louvain modularity relative errors \cite{blondel2008fast}.}
\label{tab:reconstruction_baseline_comparison}
\end{table*}

\section{Experiments}
\begin{figure*}[t]
    \centering
    \includegraphics[width=\textwidth]{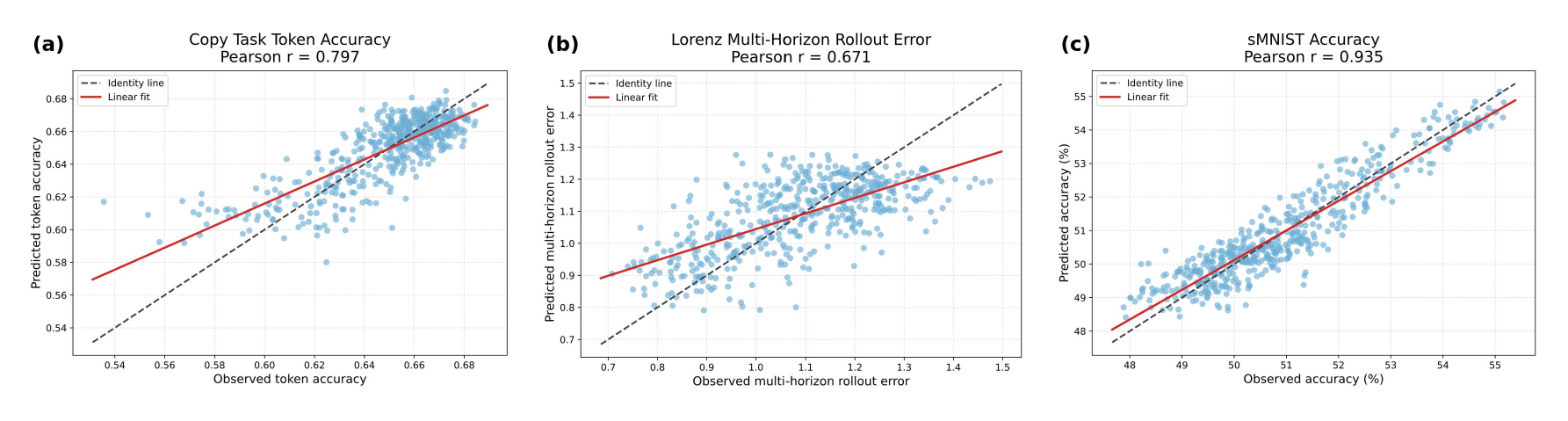}
    \caption{Predicted versus measured reservoir performance from 32-dimensional CVAE latent coordinates for copy memory, Lorenz system prediction, and sequential MNIST.}
    \label{fig:latent-to-function-predictions}
\end{figure*}

\subsection{Experimental Setup}

\paragraph{Dataset and Preprocessing} 

We utilize the IARPA MICrONS connectome dataset \cite{microns2025functional}, which comprises spatial coordinates, cell-type annotations, and single-cell synaptic connectivity mapping from mouse cortical volumes. Following prior work \cite{liu2026decodingcorticalmicrocircuitsgenerative}, we model local circuits as functional columns traversing the cortical lamina and extract fixed cylindrical subvolumes aligned orthogonally to the layer boundaries. These columnar subvolumes are arranged across the tangential $x$-$z$ plane following a hexagonal close-packing layout. Along the $z$ axis, the plane is partitioned into non-overlapping spatial regions, with the two outer regions jointly assigned to the training set and the two central regions assigned to the validation and test sets, respectively. Each sampling cylinder has a radius of $27.18~\mu\mathrm{m}$ (For reference, the overall volume spans roughly $1387~\mu\mathrm{m} \times 519~\mu\mathrm{m}$ across $x$ and $z$). To maximize data utilization and spatial coverage, neighboring sampling cylinders within the same split are allowed to overlap by approximately $30\%$ in cross-sectional area. This sampling design provides broad and approximately uniform coverage of column locations while reducing bias arising from local variations in neuronal density. For each extracted cylinder, we construct a directed subgraph containing solely the enclosed somatic nodes and their internal synaptic edges, thereby restricting the analysis to local circuit units with comparable spatial scale and orientation.

To accommodate variable neuron counts across microcircuits (100\text{--}330 nodes), each graph is padded to a uniform size of $N_{\max} = 330$ nodes. We record the number of valid nodes for each graph and mask the padded nodes during model training and evaluation. This preprocessing procedure yields the graph dataset $\{G_i\}_{i=1}^{M}$ used in the subsequent experiments, where $M = 498$.
\begin{table*}[t]
\centering
\begin{tabular}{@{}llcccccc@{}}
  \toprule
  Task & Mediator & $r_{\alpha,S}$ & $r_{S,F}$ & $\beta_{S|\alpha}$ & $\beta_{\alpha}$ & $\beta_{\alpha|S}$ & Red. \\
  \midrule
  Copy & E/I recip. & 0.787 & 0.698 & 0.267 & 0.757 & 0.547 & 0.278 \\
  Copy & CW5 & 0.527 & 0.508 & 0.151 & 0.757 & 0.678 & 0.105 \\
  Lorenz & Non-normality & -0.808 & 0.699 & 0.330 & -0.723 & -0.456 & 0.369 \\
  Lorenz & Top singular value & -0.785 & 0.649 & 0.213 & -0.723 & -0.556 & 0.231 \\
  sMNIST & Stable rank & 0.904 & 0.912 & 0.319 & 0.945 & 0.657 & 0.305 \\
  sMNIST & Top-1 energy & -0.802 & -0.825 & -0.190 & 0.945 & 0.793 & 0.161 \\
  \bottomrule
\end{tabular}
\caption{Candidate partial mediators for task-specific latent functional directions. $r_{\alpha,S}$ and $r_{S,F}$ are Pearson correlations for the two marginal associations. $\beta_{S|\alpha}$ is the mediator coefficient in the controlled model $F\sim\alpha+S$. $\beta_{\alpha}$ is the coefficient of $\alpha$ in the total-effect model $F\sim\alpha$, and $\beta_{\alpha|S}$ is the coefficient of $\alpha$ after adding the mediator in $F \sim \alpha+S$. Red. is the fractional direct-effect reduction. CW5 denotes normalized length-5 closed walks.}
\label{tab:mediation_summary}
\end{table*}

\subsection{Connectome Reconstruction and Latent Geometry} 

\paragraph{Graph-Level Fidelity}
To evaluate whether the conditional latent model recovers connectome topology at the probabilistic level, we compare it with a naive VAE baseline \cite{liu2026decodingcorticalmicrocircuitsgenerative} that excludes both the PointNet++ spatial encoder and conditional feature inputs. Edge-level reconstruction results are reported in Table~\ref{tab:reconstruction_baseline_comparison}. The full CVAE consistently outperforms the baseline in edge-probability ranking, as quantified by the area under the ROC curve (AUC). This stronger AUC suggests that the model learns an informative relative ordering over candidate edges, assigning higher probabilities to connections that are more likely under the observed connectome distribution. Such probabilistic ranking is a natural fidelity criterion for sparse cortical microcircuit graphs, where the generative task is to recover the topology of connectivity patterns rather than merely reproduce individual sampled edges.

At the graph level, the two models differ more sharply in their ability to preserve connectome topology (Figure~\ref{fig:reconstruction_latent_topology}a and Table~\ref{tab:reconstruction_baseline_comparison}). The naive baseline approximates density-sensitive statistics such as mean degree and global efficiency, but fails to preserve higher-order topological features. In contrast, the full CVAE substantially reduces relative errors in modularity, transitivity, and clustering coefficient, indicating that it retains higher-order modular and transitive structure beyond marginal edge density.

\paragraph{Latent Distributional Statistics}
We examine the learned 32-dimensional latent space with PCA while retaining the full latent vectors for downstream analyses. Across 498 connectome samples, the first two principal components explain 56.13\% of the variance. As shown in Figure~\ref{fig:reconstruction_latent_topology}b, graph-level descriptors such as clustering coefficient and mean degree change smoothly over the PC1--PC2 plane. This pattern suggests that the learned latent space preserves continuous variation in macroscopic connectome topology, even though the model is trained through node- and edge-level reconstruction objectives.

\subsection{Connectome Latents Predict Reservoir Function}

\paragraph{Downstream Tasks and Metrics}

To assess the relation between connectome latent representations and macroscopic function, we evaluate the generated connectome graphs on three complementary reservoir-computing tasks. The copy task measures delayed memory for symbolic sequences, following standard memory benchmarks in reservoir computing \cite{jaeger2002short}; Lorenz system prediction evaluates closed-loop prediction and multi-step rollout stability in a chaotic dynamical system \cite{lorenz1963deterministic}; and sequential MNIST tests the classification separability of input-driven reservoir states using the MNIST benchmark \cite{lecun1998gradient}. The functional metrics are defined in the Supplementary.  

\begin{table}[t]
\centering
\begin{tabular}{@{}lccccc@{}}
  \toprule
  Task & Linear & Ridge & SVM & XGB & GPR \\
  \midrule
  Copy & 0.624 & 0.625 & -2.477 & 0.626 & \textbf{0.651} \\
  Lorenz & 0.434 & 0.436 & 0.445 & 0.396 & \textbf{0.459} \\
  sMNIST & 0.872 & 0.872 & 0.821 & 0.844 & \textbf{0.873} \\
  \bottomrule
\end{tabular}
\caption{Latent-to-function regression performance across reservoir-computing tasks. Entries report 5-fold cross-validated $R^2$ from the 32-dimensional CVAE latent coordinates to the sample-level task metric. Higher values are better for all entries. XGB and GPR denote XGBoost and Gaussian process regression, respectively.}
\label{tab:latent_to_function_regression}
\end{table}

\paragraph{Cross-Validated Latent-to-Function Predictability}

Regression from the 32-dimensional CVAE latent coordinates to task performance shows that the learned structural representation predicts functional variation across connectome microcircuits. We compare both linear and nonlinear predictors, including Ridge regression, Gaussian process regression (GPR) \cite{rasmussen2006gaussian}, and XGBoost \cite{chen2016xgboost}, with the full comparison reported in Table~\ref{tab:latent_to_function_regression}; the corresponding predicted-versus-measured relationships are shown in Figure~\ref{fig:latent-to-function-predictions}. The strongest structure--function association appears in sequential MNIST, with a best cross-validated $R^2$ of approximately $0.87$, whereas Lorenz system prediction yields a weaker but still measurable predictive signal, with a best cross-validated $R^2$ of approximately $0.46$. Furthermore, when accounting for model complexity, simple models such as Ridge regression achieve cross-validated performance close to the best nonlinear models across tasks. This result suggests that the CVAE latent space organizes part of the graph-topological variation along functional directions that can be captured by relatively simple predictive models.

\subsection{Interpreting the Latent Space}
The goodness of fit in structure-to-function regression reveals that the latent space contains task-relevant structural variation, but it does not identify which graph properties account for this relationship. We therefore perform a mediation-style interpretability analysis to test whether task-specific structural descriptors explain part of the association between a latent functional direction and reservoir performance \cite{baron1986moderator,mackinnon2007mediation}. 

\paragraph{Gradient of Predicted Function in Latent Space}
We standardize the latent vectors across samples to obtain $\tilde{z}_i$ and fit a Ridge regression to predict the functional score:
\[
\hat{F}_i = b + \tilde{z}_i^\top w.
\]
Thus, $w$ is the gradient of the functional predictor with respect to the standardized latent coordinates: it specifies the direction in latent space along which the predicted functional score increases most rapidly. We normalize this gradient as
\[
\bar{w} = \frac{w}{\lVert w \rVert_2}
\]
and define
\[
\alpha_i = \tilde{z}_i^\top \bar{w}.
\]
Therefore, $\alpha_i$ is the signed projection of sample $i$'s latent representation onto the task-specific functional direction. It summarizes the latent variation most associated with task performance and can be directly compared with interpretable structural descriptors.

\paragraph{Understanding Latent Gradient with Graph metrics}

Next, we seek to identify specific structural descriptor $S_m$ that can explain part of the association between the task-specific latent coordinate $\alpha$ and the performance score $F$. Specifically, a descriptor $S_m$ is treated as a candidate partial mediator only if it follows the empirical pattern expected from a mediation-style mechanism: variation along $\alpha$ should be associated with variation in $S_m$; $S_m$ should be associated with task performance; and, after controlling for $S_m$, the direct association between $\alpha$ and $F$ should be reduced but not necessarily eliminated.


To assess whether a candidate descriptor satisfies these requirements, we fit four ordinary least squares regression models:
\[
S_m \sim \alpha, \qquad
F \sim S_m, \qquad
F \sim \alpha + S_m, \qquad
F \sim \alpha.
\]

We identify $S_m$ as a candidate partial mediator if it satisfies four criteria:
(1) the association $\alpha \rightarrow S_m$ is statistically significant and aligned with the hypothesized structural mechanism;
(2) $S_m \rightarrow F$ is statistically significant in the direction of superior task performance;
(3) $S_m$ retains a significant independent effect in the joint model $F \sim \alpha + S_m$ (with coefficient $\beta_{S_m \mid \alpha}$); and
(4) controlling for $S_m$ attenuates the direct effect of $\alpha$, i.e., $|\beta_{\alpha \mid S_m}| < |\beta_{\alpha}|$.
We quantify this attenuation as the fractional direct-effect reduction:
\[
\mathrm{Red} = 1 - \frac{|\beta_{\alpha \mid S_m}|}{|\beta_{\alpha}|},
\]
where $\beta_{\alpha}$ denotes the total effect of $\alpha$ in $F \sim \alpha$, and $\beta_{\alpha \mid S_m}$ is its direct effect after adjusting for $S_m$ in $F \sim \alpha + S_m$.
\subsection{Task-specific Structural Mechanisms}

Equipped with the mediation criteria defined above, we systematically evaluated candidate graph properties spanning local wiring motifs to global spectral profiles. In Table~\ref{tab:mediation_summary}, we report the most salient candidate mediators identified for each task along with their effect attenuation statistics. Across the tasks, these primary descriptors point to distinct structural mechanisms:

\begin{itemize}
    \item \textbf{Copy memory: Mixed Excitatory--Inhibitory Reciprocal Microcircuits} The copy task requires stable recurrent feedback to maintain delayed information. The strongest candidate mediator is the proportion of reciprocal connections involving mixed excitatory and inhibitory neurons. Such microcircuits provide recurrent feedback while limiting the excessive signal amplification that can arise in purely excitatory loops. Closed walks of length five provide a secondary topological mechanism, suggesting that short recurrent motifs also contribute to memory maintenance.
    \item \textbf{Lorenz system prediction: Non-normality and Dominant Singular Modes} Multi-step forecasting of Lorenz dynamics places stronger demands on dynamical stability. The latent functional direction is primarily associated with reduced matrix non-normality and a smaller dominant singular value, both of which suppress transient amplification in the reservoir dynamics. These spectral changes are consistent with improved rollout stability in chaotic prediction, where non-normal transient amplification can cause small state perturbations to grow rapidly \cite{trefethen2005spectra,hennequin2012nonnormal}.
    \item \textbf{Sequential MNIST: Singular-Spectrum Dispersion and Stable Rank} Sequential MNIST performance depends on the separability of reservoir states induced by different input classes. The main candidate mediators are the dispersion of the singular-value spectrum and the stable rank of the recurrent matrix. A more distributed singular spectrum and a higher stable rank indicate that the reservoir can support a richer set of dynamic representations, thereby improving linear separability for classification.
\end{itemize}

Taken together, these results suggest that the same connectome latent space encodes task-specific structural factors rather than a single generic notion of graph quality. Delayed memory is associated with local reciprocal microcircuits, whereas dynamical forecasting and sequential classification are more closely linked to global spectral properties of the reservoir matrix. Although the mediation analysis is observational, it provides a concrete account of how latent structural variation relates to distinct computational functions.

\section{Conclusion}

We proposed a conditional generative framework that learns compact latent representations of connectome graphs from neuronal cell type and spatial organization. Using generated connectomes as recurrent reservoirs, we found that the latent space predicts functional variation across memory, dynamic prediction, and classification. Further interpretability analysis linked task-relevant latent directions to distinct structural descriptors: mixed excitatory--inhibitory reciprocal cores for memory, and Dale-signed spectral properties for prediction and classification. These results suggest that sparse connectome graphs can be organized into a latent space that connects circuit topology with measured computational behavior.

\paragraph{Limitations}
First, all experiments are conducted on local cortical microcircuits extracted from the MICrONS dataset, and the generalizability of the learned latent space to connectomes from other brain regions, species, or spatial scales remains to be established. Second, computational function is evaluated through reservoir computing, which provides a controlled and interpretable proxy for circuit computation but does not capture the full range of biological neural dynamics or learning mechanisms. Finally, although the proposed framework generates structurally plausible connectomes and preserves structure-function relationships, the biological validity of the synthesized circuits has not been experimentally verified. Future work will investigate larger and more diverse connectome datasets, incorporate more biologically realistic dynamical models, and validate generated connectomes against additional anatomical and physiological constraints.

\section*{Acknowledgement}
This work was supported by the National Key R\&D Program of China, Project Number 2025YFA1016700. This work was also supported by the National Natural Science Foundation of China (NSFC) under Grant No. 62576011.
This work was supported in part by the Beijing Major Science and Technology Project under Contract no. Z251100008125055. This work was also supported by Beijing Academy of Artificial Intelligence (BAAI). 

\bibliography{references_arxiv}

\end{document}


\maketitle

\section{Overview}

This supplementary document provides additional implementation details,
evaluation protocols, ablations, and analyses for
\emph{Connectome-to-Function: Conditional Generative Latent Representations
for Reservoir Computing}. The main paper describes the model and summarizes
the principal findings. This document is organized to make the experiments
reproducible and to provide additional evidence for the reported
structure--function relationships.

\section{Dataset and Preprocessing}

\subsection{MICrONS Source Data and Local Microcircuit Extraction}

Local microcircuit graphs were constructed with a grid-based sampling
pipeline. In the MICrONS volume, we extracted cylindrical local regions whose
axes were aligned with the $y$ direction. Sampling centers were placed on the
$x$--$z$ plane in an approximately hexagonal pattern, which reduced uncovered
gaps and redundant overlap between neighboring candidate regions. Candidate
cylinders were constrained by \texttt{column\_ver\_num = 120} and
\texttt{min\_nodes = 100}. The validation and test partitions were defined by
spatial blocks with a split width of $2.0$ cylinder radii, limiting overlap
between neighboring samples assigned to different data splits. This procedure
produced $498$ local microcircuits, partitioned by spatial region into $415$
training graphs, $41$ validation graphs, and $42$ test graphs.

Node coordinates were normalized independently within each graph. For the
valid nodes of a graph, we subtracted the graph-wise coordinate mean and divided
by the graph-wise coordinate standard deviation. Features of padded nodes were
set to zero and excluded from model computations by the valid-node mask. Each
node was represented by a $14$-dimensional feature vector: the first $3$
dimensions contained the normalized spatial coordinates, and the remaining
$11$ dimensions encoded cell type as a one-hot vector.

\section{Conditional VAE Model Details}

\subsection{Encoder Architecture}

The encoder takes padded node features, the directed adjacency matrix, and a
valid-node mask as input. The mask is used to prevent padded nodes from
contributing to neighborhood aggregation, sequence encoding, or graph-level
pooling.

The first stage uses PointNet++ to combine spatial coordinates with cell-type
features and to extract local geometric representations. We use two set
abstraction levels. The first level samples $40$ centroids with radius $0.2$,
and the second level samples $10$ centroids with radius $0.4$. Feature
propagation then maps the hierarchical features back to the original node
resolution, yielding a geometry-aware representation for every valid node.

The second stage applies a graph attention network to incorporate local
directed connectivity. Attention is evaluated only at entries corresponding to
observed edges in the adjacency matrix, so the aggregation follows the measured
microcircuit topology rather than a fully connected node graph. The main
configuration uses three GAT layers. The numbers of attention heads are
$4$, $4$, and $16$, and the per-head hidden dimensions are $16$, $8$, and $2$,
respectively. The final concatenated node representation therefore has
$32$ dimensions.

The resulting node representations are passed to a Transformer encoder for
graph-level aggregation. A learnable \texttt{CLS} token is prepended to the node
sequence, and the Transformer encoder uses $2$ layers, $2$ attention heads, and
a feedforward dimension of $64$. The final \texttt{CLS} embedding is mapped by
two projection heads to the posterior parameters $\mu$ and $\log \sigma^2$.
A $32$-dimensional latent code $z$ is then sampled using the standard
reparameterization trick. Because spatial information has already been encoded
by PointNet++, the main configuration does not add a separate positional
encoding.

\subsection{Decoder Architecture}

The decoder generates an adjacency matrix conditioned on the latent code and a
node-level condition template. The spatial coordinates and cell-type features
of the target nodes are first processed by a PointNet++ condition encoder,
which produces a node-wise condition embedding. The $32$-dimensional latent
code is then represented as a length-one memory sequence. A Transformer decoder
uses cross-attention between the node-wise condition sequence and this memory
token, thereby incorporating the graph-level latent information into each
decoded node representation. An up-projection layer with hidden dimension
$32$ is applied before edge prediction.

Let $h_i \in \mathbb{R}^{32}$ denote the decoded representation of node $i$.
Following the adjacency convention in the main paper, $A_{ij}=1$ denotes a
directed edge from source node $j$ to target node $i$. The directed bilinear
edge head uses separate learnable projections for target and source nodes:
\[
\begin{aligned}
q_i &= W_{\mathrm{tgt}} h_i + b_{\mathrm{tgt}},\\
k_j &= W_{\mathrm{src}} h_j + b_{\mathrm{src}},\\
s_{ij} &= q_i^{\mathsf{T}} W_{\mathrm{bil}} k_j + b_{\mathrm{edge}},\\
P_{ij} &= \sigma(s_{ij}),
\end{aligned}
\]
where $s_{ij}$ is the edge score, $P_{ij}$ is the resulting edge
probability, $\sigma(\cdot)$ is the sigmoid function, and
$W_{\mathrm{tgt}}$ and $W_{\mathrm{src}}$ are independent parameter matrices.

The continuous probability matrix $P$ is converted to a binary adjacency
matrix according to the evaluation protocol used for graph reconstruction,
graph generation, or reservoir computing. When stochastic edge sampling is
specified, each candidate edge is sampled as
$A_{ij} \sim \operatorname{Bernoulli}(P_{ij})$. The diagonal is set to zero
during downstream graph construction to exclude self-loops.

\subsection{Additional Ablation Experiment}

To isolate the contribution of the spatial pathway, we perform a
PointNet++ removal ablation. The ablated model removes the PointNet++
modules from both the encoder and decoder and replaces them with node-wise
projections that preserve the corresponding interface dimensions. The dense
CVAE formulation, the full $14$-dimensional node condition, the latent
dimension, and the main training settings are otherwise kept unchanged.
Table~\ref{tab:supp_pointnet_ablation} reports edge-level reconstruction AUC
and graph-level difference ratios for the naive VAE baseline, the full model,
and the No-PointNet++ variant. The naive VAE baseline is included with the
same entries as in the main-text reconstruction table. Higher AUC indicates
better edge-probability ranking, whereas lower difference ratios indicate
closer agreement with the original graph.

\begin{table*}[t]
\centering
\begin{tabular}{@{}lcccccccc@{}}
\toprule
Model & AUC $\uparrow$ & Deg. $\downarrow$ & Eff. $\downarrow$ &
Clust. $\downarrow$ & Assort. $\downarrow$ & Mod. $\downarrow$ &
Trans. $\downarrow$ & Louv. $\downarrow$ \\
\midrule
Naive VAE baseline & 0.722 & 0.084 & 0.288 & 0.820 &
1.681 & 0.326 & 0.733 & 0.392 \\
No-PointNet++ & 0.913 & 3.106 & 1.078 & 0.196 &
3.173 & 0.485 & 0.274 & 0.421 \\
Full CVAE & 0.910 & 0.198 & 0.314 & 0.380 &
1.210 & 0.090 & 0.163 & 0.119 \\
\bottomrule
\end{tabular}
\caption{Reconstruction comparison including the naive VAE baseline,
the full CVAE, and the PointNet++ removal ablation. AUC measures edge-level
reconstruction, while Deg., Eff., Clust., Assort., Mod., Trans., and Louv.
denote the mean degree, global efficiency, clustering coefficient,
assortativity, modularity, transitivity, and directed Louvain modularity
difference ratios, respectively. Lower difference ratios indicate better
graph-level fidelity. The naive VAE baseline entries are identical to those
reported in the main paper.}
\label{tab:supp_pointnet_ablation}
\end{table*}

The full CVAE has much stronger edge-level ranking than the naive baseline and
lower errors on most higher-order graph metrics, although the baseline has
lower mean-degree and efficiency difference ratios. Within the controlled
PointNet++ ablation, the two conditional variants have comparable edge-level
ranking performance, but removing PointNet++ increases the difference ratios
for mean degree, efficiency, assortativity, modularity, transitivity, and
directed Louvain modularity; clustering coefficient is the exception. This
result separates edge-probability ranking from higher-order topological
fidelity and supports an independent role for the explicit spatial pathway in
preserving the organization of local microcircuits.

\section{Additional Figures}

\begin{figure*}[t]
\centering
\includegraphics[width=\textwidth]{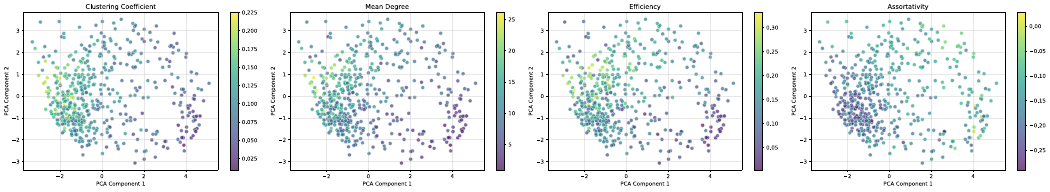}
\caption{Distribution of the learned latent representations in the first two
PCA coordinates, with points colored independently by clustering coefficient,
mean degree, efficiency, and assortativity. Each point represents one of the
$498$ connectome samples; the color bars report the corresponding graph metric.}
\label{fig:supp_latent_diagnostics}
\end{figure*}

\begin{figure*}[t]
\centering
\includegraphics[width=\textwidth]{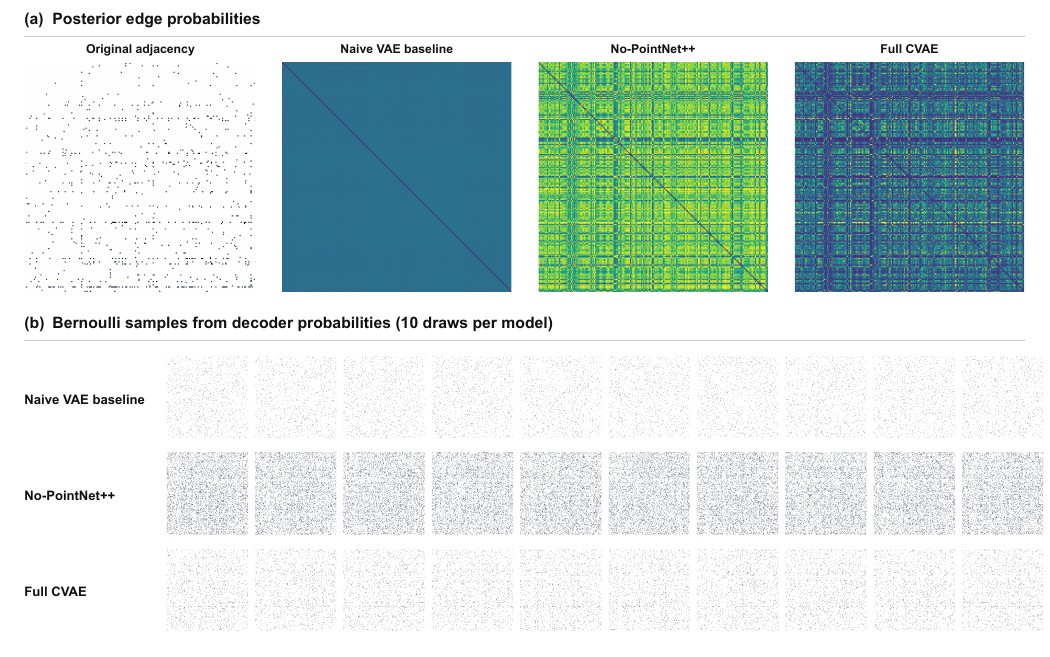}
\caption{Qualitative reconstruction comparison for one held-out connectome
sample. The top row shows the original binary adjacency matrix and the
posterior edge-probability matrices produced by the Naive VAE baseline,
No-PointNet++, and Full CVAE. The three lower rows show ten independent
Bernoulli binary samples from each model's decoder probabilities.}
\label{fig:supp_reconstruction_comparison}
\end{figure*}

Figures~\ref{fig:supp_latent_diagnostics} and
\ref{fig:supp_reconstruction_comparison} summarize the latent-space
distribution and reconstruction quality.

\section{Training Configuration}

Optimization used Adam with a batch size of $32$ for $1{,}000$ epochs, an
initial learning rate of $10^{-3}$, and a StepLR schedule with
\texttt{step\_size=500} and \texttt{gamma=0.1}. The CVAE training seed was
$42$. To reduce run-to-run numerical variation, deterministic cuDNN behavior
was enabled and benchmark mode was disabled.

The experiments ran on Ubuntu 24.04.4 LTS with NVIDIA GeForce RTX 4090 GPUs.
The software stack used Python 3.9.18 and PyTorch 2.0.1+cu118.

The KL coefficient was annealed to let the model first fit the reconstruction
term: it was set to zero for epochs $1$--$10$, increased linearly from zero to
$10^{-6}$ over epochs $11$--$60$, and then held at $10^{-6}$ through epoch
$1{,}000$.

\section{Reservoir Computing Protocols}

\subsection{Generic ESN-style Dynamics and Readout Training}

Each decoded binary adjacency matrix is instantiated as a fixed recurrent
reservoir on the valid-node support specified by the shared condition template
$X_0$. The valid-node mask of $X_0$ contains $210$ nodes, and the reservoir is
defined on this support. After masking to this support, we remove self-loops and
construct the recurrent matrix using the cell-type-dependent
excitatory or inhibitory sign of each presynaptic neuron. The recurrent matrix
$W$ and the input matrix $W_{\mathrm{in}}$ are fixed for a given connectome and
random seed, while task supervision is used to fit the linear readout.

Using the column-vector convention of the main text, the shared ESN-style
update is
\[
\begin{aligned}
a_t &= W_{\mathrm{in}}x_t + W h_{t-1},\\
h_t &= (1-lr)h_{t-1} + lr\tanh(a_t),\\
\hat{y}_t &= W_{\mathrm{out}}h_t .
\end{aligned}
\]
Here, $h_t$ is the reservoir state, $lr$ is the leak rate, and the recurrent
and input matrices are fixed after initialization. If LayerNorm is enabled in a
particular implementation, it is applied to the pre-activation across the
reservoir coordinates before the nonlinearity, with
$\tilde{a}_t=\operatorname{LN}(a_t)$ and nonlinear term
$\tanh(\tilde{a}_t)$. The normalization is non-affine and introduces no
trainable gain or bias.

The linear readout is fitted from the reservoir states. Let $H$ denote the
design matrix whose columns are reservoir states and let $Y$ denote the
corresponding target matrix. Under this column-oriented convention, ridge
readout training solves
\[
W_{\mathrm{out}}^{\star}
= YH^{\mathsf{T}}(HH^{\mathsf{T}}+\alpha I)^{-1},
\]
and the pseudoinverse solution is used when the regularization coefficient is
zero. The fitted readout remains fixed during evaluation. Functional
performance is determined by the fixed input projection, the connectome-derived
recurrent dynamics, and the information retained in the reservoir states.

\subsection{Aggregation Across Samples and Seeds}

Both decoder sampling and reservoir evaluation introduce stochastic
variation. We therefore do not use the score from a single run as the
functional label for a sample. For each latent sample $i$, the decoder
generates $10$ binary adjacency variants indexed by $a \in \{1,\ldots,10\}$.
For each fixed pair $(i,a)$, we evaluate the corresponding reservoir under
multiple random seeds indexed by $r \in \{1,\ldots,R\}$. Let
$S_{i,a,r}$ denote the task score from one such evaluation. We first average
over reservoir seeds to obtain the adjacency-level performance:
\[
\bar{S}_{i,a}
= \frac{1}{R}\sum_{r=1}^{R} S_{i,a,r}.
\]
We then average the $10$ adjacency-level scores for the same latent sample to
obtain its sample-level functional label:
\[
F_i
= \frac{1}{10}\sum_{a=1}^{10} \bar{S}_{i,a}.
\]
Thus, the averaging order separates variation due to reservoir initialization
from variation due to decoder sampling. The latent-to-function regression and
mediation-style analyses use the resulting $498$ sample-level observations
$\{F_i\}_{i=1}^{498}$.

\subsection{Downstream Task Definitions and Metrics}

\paragraph{Copy memory}

The copy-memory task requires the reservoir to reproduce a discrete token
sequence after a delay. The formal configuration uses
\texttt{T = 15}, \texttt{mem\_len = 5}, and \texttt{mem\_dim = 7}, giving a
total input sequence length of $20$. The first five positions contain randomly
sampled tokens. These are followed by the delay region; the delimiter is placed
at the sixth-to-last position, and the final five positions form the output
window. The readout target in this window is the original five-token sequence
presented at the beginning of the input.

The primary metric is token accuracy computed over the final five positions.
Let $\ell_t$ denote the readout logits at output position $t$ and let $y_t$
denote the corresponding target token. With $\mathrm{mem\_len}=5$, the metric
is
\[
\mathrm{TokenAccuracy}
= \frac{1}{5}\sum_{t=1}^{5}
\mathbf{1}\!\left[
\arg\max_{c \in \{1,\ldots,7\}} \ell_{t,c}
= y_t
\right].
\]
Higher values indicate more accurate delayed copying. Across the $498$
sample-level labels, token accuracy ranged from $0.5356$ to $0.6844$, with a
mean of $0.6486$ and a standard deviation of $0.0244$.

\paragraph{Lorenz system prediction}

The Lorenz system prediction task evaluates the ability of a reservoir to perform
multi-step closed-loop prediction in a chaotic dynamical system. We generated
the trajectory from the standard Lorenz equations,
\[
\begin{aligned}
\dot{x} &= \sigma(y-x),\\
\dot{y} &= x(\rho-z)-y,\\
\dot{z} &= xy-\beta z,
\end{aligned}
\]
using $\sigma=10$, $\rho=28$, and $\beta=8/3$. Numerical integration used a
step size of $dt=0.01$ for $20{,}000$ steps, with initial state
$(x_0,y_0,z_0)=(1,1,1)$. The resulting time series was split chronologically,
with \texttt{train\_percentage = 0.25}; no temporal shuffling was applied.

The primary metric was multi horizon rollout error. After the
reservoir was switched to closed-loop operation, prefix NRMSE was computed at
horizons
\[
\mathcal{H}=\{1,10,50,100,300,500,1000\}.
\]
For a horizon $h \in \mathcal{H}$, let $e_h$ denote the NRMSE between the
predicted and true trajectory prefixes of length $h$. The resulting
multi-horizon error curve $\{e_h\}_{h\in\mathcal{H}}$ was summarized by the
runner as a single error value. Lower values indicate more stable rollouts
across multiple prediction timescales. Across the $498$ sample-level labels,
the metric ranged from $0.707894$ to $1.47446$, with a mean of $1.08344$ and a
standard deviation of $0.154725$.

\paragraph{\sMNIST}

For sequential MNIST (\sMNIST), each $28 \times 28$ image was converted into
a sequence of length $28$ by scanning the image row by row. At each time step,
the input was a $28$-dimensional pixel vector corresponding to one image row.
The dataset was partitioned into $57{,}000$, $3{,}000$, and $10{,}000$ samples
for training, validation, and testing, respectively. The data split and sample
ordering used \texttt{data\_seed = 0}.

This metric \texttt{best\_test\_accuracy} is recorded as a
percentage, with larger values indicating better classification performance.
Across the $498$ sample-level labels, \texttt{best\_test\_accuracy} ranged from
$47.8822$ to $55.1644$, with a mean of $50.9651$ and a standard deviation of
$1.6256$.

\paragraph{Task-specific settings not stated above.}
\begin{itemize}
    \item \textbf{Copy memory:} \texttt{num\_train = 5000}, \texttt{alpha = 1e-5}, \texttt{win\_type = orthogonal}, \texttt{win\_scale = 0.5},
    \texttt{spectral\_radius = 0.999}.
    \item \textbf{Lorenz system prediction:} \texttt{units = 210}, \texttt{spectral\_radius = 0.999}, \texttt{sigma = 0.08}, \texttt{beta = 1e-4}, \texttt{seeds =
    42--46}.
    \item \textbf{sMNIST:} \texttt{units = 210}, \texttt{spectral\_radius = 0.999}, \texttt{win\_type = orthogonal}, \texttt{win\_scale = 0.3},
    \texttt{leak\_rate = 1.0}, \texttt{pooling = mean}, \texttt{alpha\_grid = \{1e-4, 1e-3, 1e-2, 0.1, 1, 10, 100\}}.
    \item \textbf{Seed aggregation:} Copy uses 3 seeds, Lorenz uses 5 seeds, and sMNIST uses 5 seeds.
\end{itemize}

\section{Additional Latent-to-Function Results}

\subsection{Regression Models}

We fit predictors from the 32-dimensional latent code to the sample-level
functional target for all $498$ samples. Copy-memory performance is measured
by token accuracy (higher is better), Lorenz performance by multi-horizon
rollout error (lower is better), and \sMNIST performance by best test accuracy
(higher is better). All models use five-fold cross-validation. Each entry in
the tables below is the mean fold-wise $R^2$ followed by its standard deviation
across folds. The cross-validated scores are the primary quantities of
interest; full-data fits are reported only as diagnostics because they can
overstate generalization.

\begin{table*}[t]
\centering
\resizebox{\textwidth}{!}{
\begin{tabular}{@{}lccccccc@{}}
\toprule
Task & Lin. & Ridge & SVM & MLP & RF & XGB & GPR \\
\midrule
Copy & $0.624 \pm 0.103$ & $0.625 \pm 0.106$ &
$-2.477 \pm 1.520$ & $-6.762 \pm 7.378$ & $\textbf{0.655} \pm 0.194$ &
$0.626 \pm 0.222$ & $0.651 \pm 0.110$ \\
Lorenz & $0.434 \pm 0.159$ & $0.436 \pm 0.161$ &
$0.445 \pm 0.125$ & $-0.012 \pm 0.331$ & $0.447 \pm 0.076$ &
$0.396 \pm 0.093$ & $\textbf{0.460} \pm 0.151$ \\
\sMNIST & $0.872 \pm 0.030$ & $0.872 \pm 0.029$ &
$0.821 \pm 0.031$ & $-9.872 \pm 2.846$ & $0.854 \pm 0.041$ &
$0.844 \pm 0.048$ & $0.873 \pm 0.030$ \\
\bottomrule
\end{tabular}
}
\caption{Five-fold cross-validated latent-to-function regression results.
Lin., RF, and XGB denote linear regression, random forest, and XGBoost,
respectively. Each entry reports mean $R^2 \pm$ fold standard deviation,
rounded to three decimals. Bold indicates the best mean score within each
task.}
\label{tab:supp_latent_regression}
\end{table*}

Across tasks, the latent space is most predictive for \sMNIST and least noisy
for the linear predictors and GPR. Copy memory and \sMNIST both show strong
performance from simple models, with GPR best or near-best in both cases. The
Lorenz task is weaker but still clearly above chance. In contrast, MLP is
unstable on all three tasks, and tree-based methods can fit the training data
well while remaining less reliable under cross-validation, so the fold-wise
$R^2$ values remain the main comparison criterion.

On the denoised Lorenz target, Random Forest gives the best cross-validated
performance at $0.4828 \pm 0.2155$, followed by GPR at $0.4672 \pm 0.1639$.
This changes the ranking slightly but not the main conclusion that the latent
representation carries usable predictive signal for the Lorenz task.

\section{Interpretability Analysis}

\subsection{Copy Memory Mechanisms}

The copy-memory target is sample-level mean token accuracy, with larger values
indicating better delayed-token recall. The expanded screen compares E/I mixed
reciprocal feedback with normalized length-5 closed walks, the two graph
descriptors retained after screening related candidates.

\begin{table*}[t]
\centering
\scriptsize
\begin{tabular}{@{}lrrrrrrl@{}}
\toprule
Candidate $S$ & $r_{\alpha,S}$ & $r_{S,F}$ & $\beta_{S|\alpha}$
& $\beta_{\alpha|S}$ & Red. & Screen \\
\midrule
E/I mixed reciprocal fraction & $0.787$ & $0.698$ & $0.267$ & $0.547$
& $27.75\%$ & Primary \\
Normalized length-5 closed walks & $0.527$ & $0.508$ & $0.151$ & $0.678$
& $10.50\%$ & Secondary \\
\bottomrule
\end{tabular}
\caption{Copy-memory mediation screen. Here $F$ denotes sample-level mean
\texttt{token\_accuracy}; $r_{\alpha,S}$ and $r_{S,F}$ are Pearson
correlations, $\beta_{S|\alpha}$ is the coefficient of $S$ in
$F\sim\alpha+S$, and $\beta_{\alpha|S}$ is the coefficient of $\alpha$ in
the same controlled model. Red. is the fractional reduction in the absolute
alpha coefficient relative to the alpha-only model. The alpha-only
$R^2=0.5735$ and $\beta_{\alpha}=0.757$; for the primary candidate,
$R^2_{\alpha+S}=0.6007$.}
\label{tab:supp_copy_mediation_screen}
\end{table*}

The expanded screen in Table~\ref{tab:supp_copy_mediation_screen} separates
one leading candidate from a weaker cycle descriptor. E/I mixed reciprocal
fraction has the largest direct-effect reduction and remains the primary
candidate. Normalized length-5 closed walks have the same direction of
association but a smaller reduction, so they are best interpreted as a
secondary partial-mediation candidate rather than an independent mechanism.

\subsection{Lorenz Forecasting Mechanisms}

The main paper highlights non-normality and the leading singular value. Here
we report the broader screen used to determine whether those descriptors were
distinct from other measures of matrix gain, conditioning, or spectral shape.

\begin{table*}[t]
\centering
\scriptsize
\begin{tabular}{@{}lrrrrrrl@{}}
\toprule
Candidate $S$ & $r_{\alpha,S}$ & $r_{S,F}$ & $\beta_{S|\alpha}$
& $\beta_{\alpha|S}$ & $R^2_{\alpha+S}$ & Red. & Screen \\
\midrule
Non-normality & $-0.8079$ & $0.6985$ & $0.330$ & $-0.456$
& $0.5602$ & $36.88\%$ & Primary \\
Condition number & $-0.7823$ & $0.6491$ & $0.216$ & $-0.554$
& $0.5404$ & $23.33\%$ & Related \\
Largest singular value & $-0.7854$ & $0.6492$ & $0.213$ & $-0.556$
& $0.5398$ & $23.13\%$ & Related \\
Effective-rank entropy & $-0.3428$ & $0.3047$ & $0.065$ & $-0.701$
& $0.5261$ & $3.06\%$ & Weak \\
Stable rank & $0.5888$ & $-0.3891$ & $0.056$ & $-0.756$
& $0.5244$ & $-4.54\%$ & Not supported \\
Spectrum slope & $0.6716$ & $-0.4366$ & $0.089$ & $-0.783$
& $0.5268$ & $-8.27\%$ & Not supported \\
\bottomrule
\end{tabular}
\caption{Expanded Lorenz mediation screen. Here $F$ denotes
\texttt{multi\_horizon\_rollout\_error}; $r_{\alpha,S}$ and $r_{S,F}$ are
Pearson correlations, $\beta_{S|\alpha}$ is the coefficient of $S$ in
$F\sim\alpha+S$, and $\beta_{\alpha|S}$ is the coefficient of $\alpha$ in
the same controlled model. Red. is the fractional reduction in the absolute
alpha coefficient relative to the alpha-only model. The alpha-only
$R^2=0.5224$ and $\beta_{\alpha}=-0.723$ are common to all rows.}
\label{tab:supp_lorenz_mediation_screen}
\end{table*}

The expanded screen in Table~\ref{tab:supp_lorenz_mediation_screen} separates
one leading candidate from several related descriptors. Non-normality has the
largest direct-effect reduction and the largest increase in explained variance.
Condition number and the leading singular value have nearly the same direction
and magnitude, so they are best interpreted as related gain/conditioning
summaries rather than three independent mechanisms. Effective-rank entropy
follows the expected direction but adds little explanatory power. Stable rank
and spectrum slope fail the direct-effect criterion, because controlling for
either one increases rather than decreases the magnitude of the alpha
coefficient.

\subsection{Sequential MNIST Mechanisms}

The \sMNIST{} target is \texttt{best\_test\_accuracy}, with larger values
indicating better classification. The expanded screen separates stable rank
from related descriptors of singular-spectrum shape.

\begin{table*}[t]
\centering
\scriptsize
\begin{tabular}{@{}lrrrrrl@{}}
\toprule
Candidate $S$ & $r_{\alpha,S}$ & $r_{S,F}$ & $R^2_{\alpha+S}$
& Red. & Screen \\
\midrule
Stable rank & $+0.9037$ & $+0.9123$ & $0.9114$ & $30.52\%$ & Primary \\
Top-1 energy fraction & $-0.8016$ & $-0.8251$ & $0.9055$ & $16.07\%$
& Related \\
Top-5 energy fraction & $-0.7573$ & $-0.7913$ & $0.9062$ & $14.24\%$
& Related \\
Spectrum slope & $+0.7803$ & $+0.7960$ & $0.9015$ & $12.40\%$
& Related \\
Effective-rank entropy & $-0.5931$ & $-0.4883$ & $0.9007$ & $-6.99\%$
& Not supported \\
\bottomrule
\end{tabular}
\caption{Expanded \sMNIST{} mediation screen for raw spectral descriptors.
Here $F$ denotes \texttt{best\_test\_accuracy}; $r_{\alpha,S}$ and $r_{S,F}$
are Pearson correlations, and Red. is the fractional reduction in the
absolute alpha coefficient relative to the alpha-only model. The alpha-only
baseline has $r_{\alpha,F}=0.9448$ and $R^2=0.8927$.}
\label{tab:supp_smnist_mediation_screen}
\end{table*}

The expanded screen in Table~\ref{tab:supp_smnist_mediation_screen} identifies
stable rank as the leading candidate and separates it from related
spectrum-shape descriptors. Stable rank has the largest direct-effect
reduction and the largest increase in explained variance. Top-1 energy
fraction, top-5 energy fraction, and spectrum slope have consistent directions
and are best interpreted as related summaries of spectral concentration rather
than three independent mechanisms. Effective-rank entropy fails the
direct-effect criterion, because controlling for it increases rather than
decreases the magnitude of the alpha coefficient.

\section{Definition of Graph Metrics}

\paragraph{Graph and preprocessing conventions.}
Each connectome sample is represented as an attributed directed graph
$G=(V,E,X)$ with a binary adjacency matrix
$A\in\{0,1\}^{N\times N}$. We use the convention from the main paper:
$A_{ij}=1$ denotes an edge from source node $j$ to target node $i$.
Padded nodes are removed using the valid-node mask, and self-loops are removed
before any binary graph metric is evaluated. Thus, all topology-only metrics
below are unweighted and loop-free. Let $\mathcal{C}$ denote the largest weakly
connected component (LWCC) of the resulting directed graph. Unless stated
otherwise, graph-level metrics are evaluated on the induced graph on
$\mathcal{C}$, with $n=|\mathcal{C}|$ nodes and adjacency matrix $A$ restricted
to these nodes. Weak connectivity is used to select the component, whereas
directed paths retain their edge orientation. Whenever a metric has a
zero-valued denominator, its value is defined as zero.

The reconstruction table uses directed binary graph metrics. In the formulas
below, $A$ is the loop-free binary adjacency matrix on the LWCC, and the
resolution parameter for modularity objectives is $\gamma=1$. The only
exception is the contextual small-worldness definition, where an undirected
support is introduced explicitly.

\paragraph{Edge-level reconstruction AUC.}
The decoder produces an edge-probability matrix
$P\in[0,1]^{N\times N}$. For sample $i$, let
\[
\mathcal{P}_i
 =
\left\{(u,v):u,v\ \text{are valid nodes},\ u\ne v\right\}
\]
be the set of valid off-diagonal node pairs, and let
$Y_{uv}=A_{uv}$ be the binary target label. The edge-level AUC is the
probability that a randomly selected positive edge receives a larger decoder
probability than a randomly selected negative pair, with ties receiving half
credit:
\[
\operatorname{AUC}_i
 =
\Pr\!\left(P_{uv}>P_{rs}\right)
 +\frac{1}{2}\Pr\!\left(P_{uv}=P_{rs}\right),
\]
where $(u,v)$ is sampled from $\mathcal{P}_i$ with $Y_{uv}=1$ and $(r,s)$ is
sampled from $\mathcal{P}_i$ with $Y_{rs}=0$. The reported AUC is the sample
mean
\[
\overline{\operatorname{AUC}}
 =
\frac{1}{M}\sum_{i=1}^{M}\operatorname{AUC}_i .
\]
The checkpoint summary records the AUC emitted by the reconstruction evaluator
before the downstream binary-graph diagonal-zeroing step; all topology metrics
below explicitly use the loop-free adjacency.

\paragraph{Relative difference ratio and aggregation.}
Let $m(G)$ be any graph metric and let $G_i^{\mathrm{real}}$ be the original
graph for sample $i$. The decoder produces ten binary adjacency variants
$G'_{i,a}$, $a=1,\ldots,10$. We first compute the metric for each variant and
average at the sample level:
\[
\overline{m}_{i}^{\,\mathrm{gen}}
 =
\frac{1}{10}\sum_{a=1}^{10}m\!\left(G'_{i,a}\right).
\]
The sample-level relative difference ratio is
\[
\operatorname{DR}_{i,m}
 =
\frac{
\left|m\!\left(G_i^{\mathrm{real}}\right)
-\overline{m}_{i}^{\,\mathrm{gen}}\right|
}{
\left|m\!\left(G_i^{\mathrm{real}}\right)\right|+\epsilon
},
\qquad
\epsilon=10^{-8}.
\]
Finally, the value reported in the reconstruction tables is
\[
\overline{\operatorname{DR}}_{m}
 =
\frac{1}{M}\sum_{i=1}^{M}\operatorname{DR}_{i,m}.
\]
Thus, variant averaging precedes the relative-difference calculation, and
sample averaging is performed last. Smaller values indicate closer agreement
with the original graph. Because the denominator uses the absolute value of
the original metric, ratios can become large when that metric is close to
zero; this is particularly relevant for degree assortativity.

\paragraph{Degree and density.}
Let $m_{\mathrm{d}}=|E|=\sum_{i,j}A_{ij}$ be the number of directed edges.
For node $i$, the in-degree, out-degree, and total degree are
\[
k_i^{\mathrm{in}}=\sum_j A_{ij},\qquad
k_i^{\mathrm{out}}=\sum_j A_{ji},\qquad
k_i=k_i^{\mathrm{in}}+k_i^{\mathrm{out}}.
\]
The mean degree used in the reconstruction table is the mean total directed
degree,
\[
\overline{k}
 =
\frac{1}{n}\sum_{i=1}^{n}k_i
 =
\frac{2m_{\mathrm{d}}}{n}.
\]
For reference, the directed edge density is
\[
\rho
 =
\frac{m_{\mathrm{d}}}{n(n-1)}.
\]
Density is used as a contextual density-sensitive descriptor but is not one of
the seven relative-error columns in the main reconstruction table.

\paragraph{Global efficiency.}
Let $d_{ij}$ be the length of the shortest directed path from node $i$ to
node $j$, with $d_{ij}=\infty$ when $j$ is unreachable from $i$. Global
efficiency is the mean reciprocal shortest-path distance,
\[
E_{\mathrm{glob}}
 =
\frac{1}{n(n-1)}
\sum_{\substack{i,j=1\\i\ne j}}^{n}
\frac{1}{d_{ij}},
\qquad
\frac{1}{\infty}=0.
\]
The LWCC restriction removes disconnected components at the graph-selection
stage, but directed unreachable pairs within the LWCC still contribute zero.

\paragraph{Directed clustering coefficient.}
We use the directed binary clustering coefficient. Let
\[
\begin{aligned}
S&=A+A^{\mathsf T},\\
\tau_i&=\frac{1}{2}(S^3)_{ii},\\
\omega_i&=k_i(k_i-1)-2(A^2)_{ii}.
\end{aligned}
\]
Here $\tau_i$ counts directed triangle patterns around node $i$, while
$\omega_i$ is the number of possible directed triangle patterns after
subtracting reciprocal two-edge pairs that cannot form a third-node triangle.
The local coefficient and graph-level mean are
\[
c_i=
\begin{cases}
\displaystyle\frac{\tau_i}{\omega_i},&\omega_i>0,\\[6pt]
0,&\omega_i=0,
\end{cases}
\qquad
C_{\mathrm{dir}}=\frac{1}{n}\sum_{i=1}^{n}c_i.
\]

\paragraph{Directed degree assortativity.}
Degree assortativity is computed with the directed binary out--in convention.
For each directed edge from source $j$ to target $i$, define
$x_{ij}=k_i^{\mathrm{in}}$ and $y_{ij}=k_j^{\mathrm{out}}$. With
$m_{\mathrm{d}}=\sum_{i,j}A_{ij}$,
\[
\begin{aligned}
\mu_{1}
&=
\frac{1}{m_{\mathrm{d}}}
\sum_{i,j:A_{ij}=1}
\frac{x_{ij}+y_{ij}}{2},\\
\mu_{2}
&=
\frac{1}{m_{\mathrm{d}}}
\sum_{i,j:A_{ij}=1}
\frac{x_{ij}^2+y_{ij}^2}{2}.
\end{aligned}
\]
The assortativity coefficient is
\[
\begin{aligned}
r_{\mathrm{dir}}
 =
\frac{
\displaystyle
\frac{1}{m_{\mathrm{d}}}\sum_{i,j:A_{ij}=1}x_{ij}y_{ij}
-\mu_1^2
}{
\mu_2-\mu_1^2
}.
\end{aligned}
\]
Positive values indicate that edges tend to connect high out-degree sources to
high in-degree targets, whereas negative values indicate disassortative
source--target degree matching.

\paragraph{Directed modularity.}
Let $c_i$ be the community label of node $i$. The directed modularity objective
is
\[
Q_{\mathrm{dir}}(A,c)
 =
\frac{1}{m_{\mathrm{d}}}
\sum_{i,j}
\left(
A_{ij}-\frac{k_i^{\mathrm{in}}k_j^{\mathrm{out}}}{m_{\mathrm{d}}}
\right)
\mathbf{1}[c_i=c_j].
\]
The null-model term follows the adjacency convention: $i$ is the target and
$j$ is the source. The modularity column in the reconstruction table reports
\[
Q_{\mathrm{mod}}=Q_{\mathrm{dir}}(A,c^{\mathrm{spec}}),
\]
where $c^{\mathrm{spec}}$ is the partition selected by the deterministic
directed modularity optimization routine.

\paragraph{Directed transitivity.}
Global transitivity uses the same directed binary triangle counts as
$C_{\mathrm{dir}}$, but pools numerator and denominator before forming the
ratio:
\[
T_{\mathrm{dir}}
 =
\begin{cases}
\displaystyle
\frac{\sum_i \tau_i}{\sum_i \omega_i},
& \sum_i \omega_i>0,\\[6pt]
0,& \sum_i \omega_i=0.
\end{cases}
\]
Unlike $C_{\mathrm{dir}}$, which gives every node equal weight, transitivity
weights nodes according to the number of directed triples they contribute.

\paragraph{Directed Louvain modularity.}
Directed Louvain modularity uses the same $Q_{\mathrm{dir}}$ objective, but
selects the partition with the directed Louvain heuristic:
\[
c^{\mathrm{Louvain}}\approx \arg\max_c Q_{\mathrm{dir}}(A,c).
\]
The reported statistic is
\[
Q_{\mathrm{Louvain}}^{\mathrm{dir}}
 =
Q_{\mathrm{dir}}\!\left(A,c^{\mathrm{Louvain}}\right).
\]
This metric is kept separate from $Q_{\mathrm{mod}}$ because the partition is
obtained by a Louvain heuristic rather than by the deterministic spectral
modularity routine.

\paragraph{Small-worldness.}
Small-worldness is mentioned in the main paper as a standard contextual
connectome descriptor. For this contextual metric only, let
\[
B_{ij}=\mathbf{1}\!\left[A_{ij}+A_{ji}>0\right],\qquad B_{ii}=0
\]
be the undirected support. Using the undirected clustering coefficient
$C^{(u)}$ and mean shortest-path length $L^{(u)}$ on $B$, and denoting the
corresponding values from a degree-matched random-graph ensemble by
$C_{\mathrm{null}}$ and $L_{\mathrm{null}}$, the normalized small-worldness
index is
\[
\sigma_{\mathrm{SW}}
 =
\frac{C^{(u)}/C_{\mathrm{null}}}{L^{(u)}/L_{\mathrm{null}}}.
\]
Values above one indicate greater clustering relative to the increase in path
length compared with the null ensemble. This contextual index is not a
separate column in the reconstruction table.

\paragraph{E/I mixed reciprocal fraction.}
Let $s_i\in\{+1,-1\}$ denote the excitatory or inhibitory identity of neuron
$i$. The set of reciprocal unordered pairs is
\[
\mathcal{R}
 =
\left\{\{i,j\}:i<j,\ A_{ij}A_{ji}=1\right\},
\]
and its mixed E/I subset is
\[
\mathcal{R}_{\mathrm{EI}}
 =
\left\{\{i,j\}\in\mathcal{R}:s_i\ne s_j\right\}.
\]
The reciprocal E/I fraction used in the copy-memory analysis is
\[
f_{\mathrm{EI}}^{\mathrm{recip}}
 =
\begin{cases}
\displaystyle\frac{|\mathcal{R}_{\mathrm{EI}}|}{|\mathcal{R}|},
&|\mathcal{R}|>0,\\[6pt]
0,&|\mathcal{R}|=0.
\end{cases}
\]
Each unordered reciprocal pair is counted once, regardless of which direction
is used as the source in the adjacency matrix.

\paragraph{Normalized length-5 closed walks.}
The number of directed closed walks of length five, with a distinguished
starting node, is
\[
W_5
 =
\operatorname{tr}(A^5)
 =
\sum_{i_1,\ldots,i_5}
A_{i_1i_2}A_{i_2i_3}A_{i_3i_4}A_{i_4i_5}A_{i_5i_1}.
\]
The normalized length-5 closed-walk descriptor is node-normalized:
\[
\operatorname{CW5}_{\mathrm{norm}}
 =
\frac{\operatorname{tr}(A^5)}{n}.
\]
This normalization compares the average number of length-5 closed walks per
node and does not identify cyclic walks that differ only by their starting
point as a single occurrence.

\paragraph{Recurrent matrix and singular-spectrum metrics.}

For a valid connectome graph, let $s_j\in\{-1,+1\}$ be the Dale sign of the
presynaptic neuron $j$. The recurrent matrix used for the spectral descriptors
is
\[
W_{ij}=A_{ij}s_j.
\]
Let
\[
W=U\operatorname{diag}(\sigma_1,\ldots,\sigma_r)V^{\mathsf T},
\qquad
\sigma_1\ge\cdots\ge\sigma_r\ge 0
\]
be a singular-value decomposition, and let
$\|W\|_F^2=\sum_{\ell=1}^{r}\sigma_\ell^2$.
The non-normality, largest singular value, stable rank, and top-1 singular
energy fraction are, respectively,
\[
\mathcal{N}(W)
 =
\left\|W^{\mathsf T}W-WW^{\mathsf T}\right\|_F,
\]
\[
\sigma_{\max}(W)=\sigma_1(W)=\|W\|_2,
\]
\[
\operatorname{srank}(W)
 =
\frac{\|W\|_F^2}{\|W\|_2^2}
 =
\frac{\sum_{\ell=1}^{r}\sigma_\ell^2}{\sigma_1^2},
\]
and
\[
\eta_1(W)
 =
\frac{\sigma_1^2}{\sum_{\ell=1}^{r}\sigma_\ell^2}.
\]
For the zero matrix, the stable rank and $\eta_1$ are defined as zero. A
spectral-radius rescaling, when required by a reservoir task, is a separate
dynamical preprocessing step; the definitions above refer to the
matrix whose spectrum is being analyzed.